# Türkçe Konuşmadan Metne Dönüştürme için Ön Eğitimli Modellerin Performans Karşılaştırması: Whisper-Small ve Wav2Vec2-XLS-R-300M

# Performance Comparison of Pre-trained Models for Speech-to-Text in Turkish: Whisper-Small and Wav2Vec2-XLS-R-300M

Öykü Berfin MERCAN
*AdresGezgini A.Ş.*
*Ar-Ge Departmanı*
*İzmir, Türkiye*
oykumercan@adresgezgini.com
ORCID: 0000-0001-7727-0197

Sercan ÇEPNİ
*AdresGezgini A.Ş.*
*Ar-Ge Departmanı*
*İzmir, Türkiye*
sercancepni@adresgezgini.com
ORCID: 0000-0002-3405-6059

Davut Emre TAŞAR
*Dokuz Eylül Üniversitesi*
*Yönetim Bilişim Sistemleri*
*İzmir, Türkiye*
davutemre.tasar@ogr.deu.edu.tr
ORCID:0000-0002-7788-0478

Şükrü OZAN
*AdresGezgini A.Ş.*
*Ar-Ge Departmanı*
*İzmir, Türkiye*
sukruozan@adresgezgini.com
ORCID: 0000-0002-3227-348X

## Öz

*Bu çalışmada konuşmadan metne çeviri için önerilmiş ve çok sayıda dille ön eğitilmiş iki model olan Whisper-Small ve Wav2Vec2-XLS-R-300M modellerinin Türkçe dilinde konuşmadan metne çevirme başarıları incelenmiştir. Çalışmada açık kaynaklı bir veri kümesi olan Türkçe dilinde hazırlanmış Mozilla Common Voice 11.0 versiyonu kullanılmıştır. Az sayıda veri içeren bu veri kümesi ile çok dilli modeller olan Whisper-Small ve Wav2Vec2-XLS-R-300M ince ayar (fine-tune) yapılmıştır. İki modelin konuşmadan metne çeviri performansı değerlendirilmiş ve Wav2Vec2-XLS-R-300M modelinin 0,28 WER değeri Whisper-Small modelinin 0,16 WER değeri gösterdiği gözlemlenmiştir. Ek olarak modellerin başarısı eğitim ve doğrulama veri kümesinde bulunmayan çağrı merkezi kayıtlarıyla hazırlanmış test verisiyle incelenmiştir.*

## Abstract

*In this study, the performances of the Whisper-Small and Wav2Vec2-XLS-R-300M models which are two pre-trained multilingual models for speech to text were examined for the Turkish language. Mozilla Common Voice version 11.0 which is prepared in Turkish language and is an open-source data set, was used in the study. The multilingual models, Whisper-Small and Wav2Vec2-XLS-R-300M were fine-tuned with this data set which contains a small amount of data. The speech to text performance of the two models was compared. WER values are calculated as 0.28 and 0.16 for the Wav2Vec2-XLS-R-300M and the Whisper-Small models respectively. In addition, the performances of the models were examined with the test data prepared with call center records that were not included in the training and validation dataset.*

**Keywords***:* Speech to Text, Whisper, Wav2Vec2-XLS-R, WER, Common Voice, Fine-tuning

## 1. Giriş

Konuşmadan Metne Dönüştürme (Speech To Text (STT)), Doğal Dil İşleme'de (Natural Language Processing (NLP)) aktif bir araştırma alanıdır. İnsan-makine etkileşimini amaçlayan sesten metne dönüştürme alanına ise yoğun ilgi gösterilmektedir. Özellikle pandemi süreciyle birlikte evden çalışma kültürünün yaygınlaşmasıyla mesai saatleri içerisinde iş ve eğitim başta olmak üzere iletişim kurmak amacıyla mesajlaşma ve çevrimiçi olarak sesli yada video görüşmeleri gerçekleştirilmektedir. Gerçekleşen görüşmeler esnasında konuşulanlar not edilmediğinde yada kayıt altına alınmadığında tekrar erişmek mümkün olmamaktadır. Görüşme kayıtlarının tekrar tekrar incelenmesi toplantı süresi kadar bir zaman kaybı yaratarak verimsiz bir yöntem olmakla birlikte görüşme sırasında katılımcılar tarafından not alınsa dahi bu durum kişinin not alma becerisi ve dikkati ile kısıtlıdır. Çevrimiçi toplantılar haricinde halihazırda müşteri ilişkilerinde aktif olarak çağrı merkezi kullanan firmalar için hizmet ve kalite düzeyi takibi amacıyla görüşmelerin kalite yöneticisi tarafında kontrolünü sağlamak için konuşmadan metne çeviri önem arz etmektedir.

Bu çalışma, Türkçe Konuşmadan Metne Dönüştürme (Speech To Text, STT) alanında son teknoloji modellerin performanslarının sistematik bir karşılaştırılmasını sunarak literatüre önemli katkılar sağlamaktadır. Özellikle, literatürde yer alan popüler, derin öğrenme tabanlı iki farklı modelin Kelime Hata Oranı (Word Error Rate, WER) performansını Türkçe konuşma tanıma bağlamında değerlendiren bu çalışma, model seçimi ve uygulama süreçlerinde araştırmacılar ve uygulayıcılar için önemli bir rehber niteliği taşımaktadır.

## 2. Benzer Çalışmalar

Literatürde, farklı STT model yaklaşımlarını öneren çalışmalar yapılmıştır. Örneğin, araştırmacılar, telefon görüşmelerinin metin şeklinde depolanmasını sağlamak, farklı dillerdeki konuşmaları yazıya dökmek ve çevirmek, işitme engelli kişiler için ses içeriğine erişilebilirliği artırmak, konuşmayı gerçek zamanlı olarak yazıya dökmek, analiz etmek ve konuşmaya dayalı sanal asistanlar geliştirmek için STT modellerini kullanmışlardır [1, 2 ,3]. Aktif olarak telefon görüşmelerinin gerçekleştiği çağrı merkezlerinde, görüşmelerdeki sahtekarlık tespiti üzerine odaklanmıştır. Çağrı merkezi kayıtları ilk olarak konuşma tanıma sistemine verilip konuşmalar metinlere dönüştürülmüş ardından metin sınıflandırma gerçekleştirilerek sahtekarlık tespiti gerçekleştiren sistem geliştirilmiştir [1]. Konuşmayı Hint işaret diline çeviren otomatik bir sistem geliştirerek işitme engelliler ile işiten insanlar arasında iletişim kopukluğuna çözüm getirilmesi amaçlanmıştır [2]. Bir diğer çalışmada, ALS kaynaklı dizartrisi olan kişilerin, yaygın olarak kullanılan üç akıllı telefon tabanlı asistanlar tarafından ne ölçüde anlaşıldığı ve tutarlı cevaplar alabildiği araştırılmış ve özellikle İtalyanca dizartrik konuşmanın tanınması üzerine odaklanılmıştır [3]. Bu çalışmalar, STT modellerinin çeşitli alanlardaki potansiyelini göstermiş ve bu alanda sürekli araştırma ve geliştirmenin önemini vurgulamıştır [4]. Bu pratik uygulamalara ek olarak, STT modellerinin doğruluğunu ve hızını geliştirmeye yönelik çalışmalar da yapılmıştır. Araştırmacılar, evrişimli sinir ağları (CNN'ler), tekrarlayan sinir ağları (RNN'ler) ve dönüştürücüler dahil olmak üzere çeşitli derin öğrenme mimarilerini araştırmış ve STT modellerinin performansını artırmak için transfer öğrenme ve çok görevli öğrenme gibi yeni eğitim yaklaşımları geliştirmişlerdir [5].

STT alanındaki bir başka aktif çalışma alanı da gürültüyü, aksanları ve konuşmadaki diğer varyasyonları işleyebilen modellerin geliştirilmesidir. Bu modeller, STT sistemlerinin başarısını artırmayı ve konuşma kayıtlarının düşük kalitede olabileceği pratikteki uygulamalara uygun hale getirmek üzere çalışmaları içermektedir [6]. STT teknolojisinin geliştirilmesinde dikkate alınması gereken bir diğer önemli husus gizlilik ve güvenliktir. Konuşma verileri son derece kişisel ve hassas olduğundan, STT sistemlerinin gizlilik ve güvenlik göz önünde bulundurularak tasarlandığından emin olmak önemlidir. Araştırmacılar, bu endişeleri gidermek için gizliliği koruyan makine öğrenimi ve birleşik öğrenme gibi çeşitli teknikleri araştırmaktadır [7].

Dikkat tabanlı kodlayıcı ve kod çözücü modellemesi konuşmadan metne görevi için önemli gelişmeler sağlamıştır. Dikkat tabanlı tekrarlayan sinir ağlarını konuşma tanıma için uygulanabilir kılan eklentiler sunulmuştur [8]. Üretilen her fonem için, bir dikkat mekanizması, giriş dizisindeki potansiyel olarak tüm zaman adımlarında eğitilmiş bir öznitelik çıkarma mekanizması tarafından üretilen sinyalleri seçer veya ağırlıklandırır. Ağırlıklı özellik vektörü daha sonra çıktı dizisinin bir sonraki öğesinin oluşturulmasını koşullandırmaya yardımcı olur. Bir başka çalışmada ağ bileşenlerinin ön eğitiminin etkisini gösterilmiş ve kodlayıcı ile kod çözücü arasına ek bir katman ekleyerek verimli bir şekilde nasıl birleştirileceği araştırılmıştır. Ekstra katmanın daha iyi ortak öğrenme sağladığı ve performans artışı gösterdiği gözlemlenmiştir [9]. [10]'de otomatik konuşma tanıma ve konuşma çevirisi görevleri için metin verilerinden yararlanmak üzere genel bir çoklu görev öğrenme çerçevesi sunulmuştur. Otomatik konuşma tanıma bir gürültü giderici otomatik kodlayıcı göreviyle birlikte eğitilirken, makine çevirisi görevi konuşma çevirisi göreviyle paralel verilerle ortaklaşa eğitilmiştir.

Literatürde yer alan Türkçe dili özelinde gerçekleştirilen konuşmadan metne çeviri çalışmaları araştırılmış, son yıllarda kaydedilen ilerlemeler ve performans sonuçları incelenmiştir. Çalışmalarda Türkçenin foneme dayalı morfolojik yapısı dikkate alınmıştır. Destek Vektör Makineleri (SVM) tabanlı Türkçe konuşmayı metne dönüştürme sistemi geliştirilmiştir [11]. Türkçe konuşmanın özelliklerini çıkarmak için Mel Frequency Cepstral Katsayıları (MFCC) uygulanmış ve fonemleri sınıflandırmak için SVM tabanlı sınıflandırıcı kullanılmıştır. Başka bir çalışmada geleneksel Gauss karışım model-saklı Markov modelinden daha iyi sonuçlar veren mobil kayıtlı bir Türkçe veri seti üzerinde Türkçe dilinde konuşmadan metne çeviri için derin sinir ağı modeli önerilmiştir [12]. Yine benzer bir çalışmada derin öğrenme tabanlı konuşmadan metne çevirinin Gauss karışım model-saklı Markov modeline göre başarılı sonuçlar vermiş olup çalışma [12]'de sunulan çalışma ile benzerlik göstermektedir [13]. [12]'den farklı olarak [13]'de farklı eğitim verisi ve ileri beslemeli sinir ağı yerine derin inanç ağları kullanılmıştır. [14]'de LSTM ve GRU tabanlı sistemler ile geleneksel yöntemler ve derin sinir ağı tabanlı sistemlerin konuşma tanıma performansı karşılaştırılmıştır. Konuşma tanıma çalışmalarının artmasıyla birlikte etiketli veri sayısının az olması temel problem haline gelmiştir. Bu sorunun çözümü için daha az etiketli veri gerektiren denetimsiz öğrenme yaklaşımları önerilmiştir. [15]'de YouTube'dan seçilen 6,5 bin saatlik etiketsiz ses verisi ile ön eğitimli kendi kendine denetimli öğrenmenin avantajlarından yararlanarak Türkçe dilinde konuşma tanıma yaklaşımı geliştirmişlerdir.

Sınırlı kaynaklarla son teknoloji bir konuşma çevirisi sistemi oluşturmak için [16]'de önceden eğitilmiş konuşma tanıma ve metin çevirisi modelleri kullanılması önerilmiştir. Teknolojinin ilerlemesi ve konuşma tabanlı uygulamalara yönelik artan talep, STT'yi konuşma tanıma, transkripsiyon ve analiz için yaygın olarak kullanılan bir araç haline getirmiştir [17]. Wav2Vec2.0, HuBERT gibi kendi kendini denetleyen ses kodlayıcılar, yüksek kaliteli ses temsillerini öğrenir [15, 18]. Bu modeller denetimsiz ön eğitim yapısı nedeniyle, ses temsillerini kullanılabilir çıktılara dönüştürmek için uygun bir kod çözücü yapısına ihtiyaç duyarlar. Sesten metne çeviri görevi için modellerin ince ayarı gerekmektedir. Wav2vec2.0 [17] gibi kendi kendini denetleyen derin öğrenme modelleri, STT alanında dikkate değer sonuçlar göstermiştir. 2020'de Facebook AI Research tarafından geliştirilen wav2vec2.0 modeli, çok büyük bir konuşma verisi külliyatı üzerinde eğitilmiş transformer tabanlı bir model olarak sunulmuştur. Öte yandan, OpenAI tarafından 2021'de [19] tanıtılan Whisper, derin sinir ağı tabanlı bir mimari kullanan ve daha farklı bir yaklaşım kullanılarak eğitilmiş, gerçek dünya uygulamaları için uygun hale getirilmiş daha yeni bir STT modeli olarak sunulmuştur. [18]'de nöral tabanlı konuşmadan metne çeviri mimarilerinden Wav2Vec2.0 ve Whisper'ın ince ayarlı bir versiyonları üzerinde performans karşılaştırması yapılmış ve bu modeller İngilizce, İspanyolca, Almanca, Fransızca, İtalyanca, Portekizce ve Lehçe dahil olmak üzere yedi Hint-Avrupa dilinden alınan veriler üzerinde test edilmiştir.

Karşılaştırmalı bir analiz yapabilmek için Türkçe dilinde otomatik konuşma tanıma görevinde yapılmış olan ince ayar çalışmaları araştırılmıştır. Bir yüksek lisans tezi olarak yapılan çalışmada [20], Mozilla Common Voice veri setinin daha eski bir versiyonu olan 6.1 versiyonu ile bir wav2vec2.0 modeline ince ayar çalışması yapılmış ve 0,23'lük bir kelime hata oranı elde edilmiştir. Bir başka çalışmada [21] ise yine wav2vec2-XLSR-53 modelinin Common Voice, Youtube ve Ses kayıtları kullanılarak ince ayar yapıldığı görülmüştür. Whisper ile ilgili ise, Türkçe dili için herhangi bir ince ayar çalışması ile karşılaşılmamıştır. Tespit edilmiş olan bu çalışmalara ait sonuçlar ise aşağıdaki Çizelge-1'de gösterilmiştir.

**Çizelge-1: Benzer çalışmalar**

| Model | Veri Seti | WER |
|---|---|---|
| Wav2Vec2.0-XLSR [20] | Common Voice 6.1 | 0,23 |
| Wav2Vec2.0-XLSR-53 [21] | Common Voice<br>Youtube<br>Ses Kayıtları | 0,32 |

Bu çalışmada ise son teknoloji, derin öğrenme tabanlı iki model olan wav2vec2.0 [17] ve Whisper'ın [19] Kelime Hata Oranı (Word Error Rate (WER)) performansları Türkçe konuşma tanıma bağlamında karşılaştırılmıştır. Türkçe konuşma tanımada wav2vec2.0 ve Whisper'ın WER performansını karşılaştırmak için Mozilla Common Voice veri kümesi [22] kullanılmıştır. Mozilla Common Voice veri kümesi, Türkçe dahil birçok dilde ses kaydı içeren, STT modelleri için yaygın olarak kullanılan bir veri kümesidir. Deneysel sonuçlar, bu makalenin aşağıdaki bölümlerinde ayrıntılı olarak sunulmaktadır.

Türkçe dahil olmak üzere birçok dille ön eğitimli olan iki modelin karşılaştırmasının sunulduğu bu çalışma, Türkçe STT alanındaki teknoloji ve metodolojinin mevcut durumunun kapsamlı bir değerlendirmesini sağlamaktadır. Önerilen çalışmanın sonuçları, özellikle Türkçe konuşma tanıma bağlamında, NLP ve STT alanındaki araştırmacılar ve uygulayıcılar için doğru modeli seçme konusunda fikir verecek bir çıktı içermektedir.

## 3. Tasarım ve Yöntem

Bu çalışmada, Türkçe dilinde konuşmadan metne çevirme amaçlanmış olup önerilen iki farklı modelin başarısı değerlendirilmiştir. Modellerin ince ayarı için açık kaynaklı veri kümesi Common Voice veri kümesi kullanmıştır. Bu bölümde ise çalışmada kullanılan veri kümesi, veri kümesine uygulanan ön işlemler ve konuşmadan metne çeviri modelleri için ince ayar gerçekleştirilecek modeller açıklanacaktır. Deneysel akış diyagramı Şekil-1'de verilmiştir.

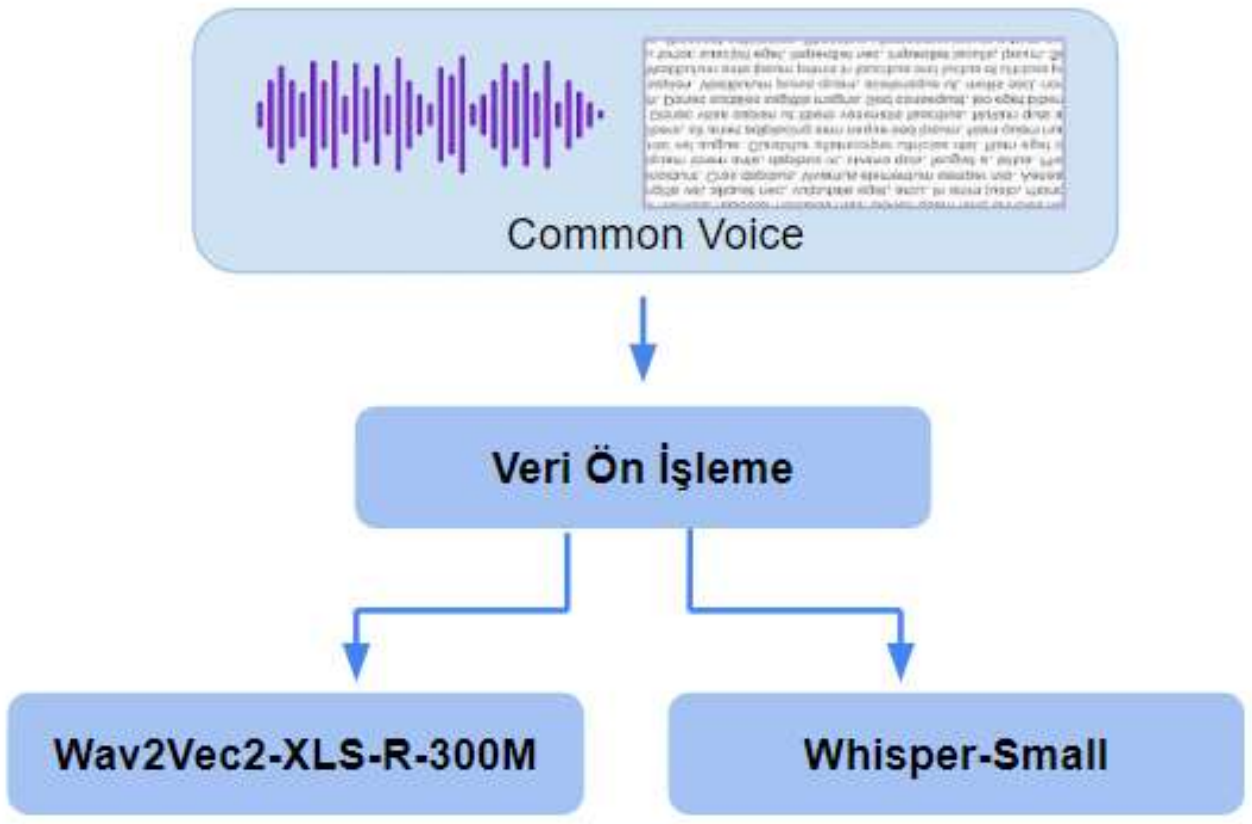

**Şekil-1:** Akış diyagramı

## 3.1 Veri Kümesi

Konuşmadan metne çeviri modellerinin düşük kelime hata oranlarını ile başarılı sonuçlar üretebilmesi için çok sayıda konuşma-metin çiftinden oluşan veri kümesi ile eğitilmesi gerekmektedir. Mozilla Common Voice, ses teknoloji üzerine çalışan geliştiricileri destekleyici çok sayıda ses kaydı biriktirerek derin öğrenme mimarilerinin eğitimi için önemli ses kaynağı olma niteliği taşımaktadır. İlk veri kümesi 2017 yılında 500 saatlik İngilizce cümlelerin seslendirilmesi ile oluşturulan Common Voice veri tabanı 2022 yılı itibariyle 100 farklı dilde ses kaydı içermektedir. Önerilen çalışmada Türkçe dilinde sesten metne çeviri gerçekleştirmek amaçlanmış olup model eğitimleri Türkçe dilinde hazırlanmış Common Voice Corpus 11.0 veri kümesi ile gerçekleştirilmiştir. Veri kümesi, 1.82 GB boyutunda ve 1328 kişi tarafında seslendirilmiş 88 saatlik kayıt içermekte olup ses kayıtları MP3 (Motion Pictures Experts Group Layer 3) formatında kaydedilmiştir. Ses kayıtları ve bu ses kayıtlarıyla eşleşen metinler konuşmadan metne çeviri modellerine girdi verisi oluşturacak şekilde ön işlemlerden geçirilmiştir.

## 3.2 Veri Ön İşleme

Model eğitimi ve doğrulama aşamasında kullanılacak veri kümesi ilk olarak bir dizi ön işlemden geçirilmiştir. Common Voice veri kümesinde ses kayıtlarına denk gelen metinler, büyük-küçük harf kullanımı ve noktalama işaretleri içermektedir. Cümle içinde yer alan bu noktalama işaretleri konuşma içerisinde herhangi bir ses birimine karşılık gelmemektedir. Bu nedenle ses parçasını noktalama işaretine göre sınıflandırmak mümkün olmamaktadır. Bu durum göz önüne alınarak kelime anlamına katkısı olmayan ve bir sesle temsil edilmeyen noktalama işaretleri veri kümesinden kaldırılmıştır. Bir diğer düzenleme ise büyük harfler üzerinde gerçekleştirilmiştir. Veri kümesinde bulunan her harf için bir belirteç atanmaktadır. Aynı harfin büyük ve küçük halinin veri kümesinde bulunması o harfe ait iki belirteç tanımlanmasına neden olmaktadır. Veri kümesindeki tüm harfler küçük harfe dönüştürülerek normalize edilmiştir. Son olarak karakter çeşitliliğini azaltmak amacıyla kullanımı yaygın olmayan şapkalı harfler (â, î, ô, û) dönüştürülerek veri kümesi sadeleştirilmiştir. Whisper model eğitimi için sadece şapkalı harflerde düzenleme yapılmıştır. Bunun nedeni Whisper sesten metne dönüştürme işleminin yanında imla kurallarına uygun bir şekilde sonuç vermektedir.

## 3.3 Yöntem

Çalışmada, konuşmadan metne çeviri için Facebook AI ve OpenAI tarafından geliştirilmiş, birçok farklı dilde ön eğitimli iki model olan Wav2Vec2-XLS-R-300M ve Whisper-Small modelleri önerilmiştir. Bu başlık altında önerilen bu modeller detaylı olarak açıklanacaktır.

### 3.3.1 Wav2Vec2-XLS-R-300M

2020 yılında Facebook AI denetimsiz ön eğitim tekniği ile geliştirdikleri STT modeli Wav2Vec2.0'i önermişlerdir [9]. Önerilen bu yöntem ile model, etiketli veriye ihtiyaç duymadan çok sayıda ham ses verisiyle ön eğitilir. Wav2Vec2.0 ardından Conneau vd. 53 farklı dilde ön eğitimli çok dilli bir model olan Wav2Vec2-XLS-R-53 modelini önermişlerdir [23]. XLSR modeli 8 dil içeren 50,7 bin saatlik çok dilli LibriSpeech, 36 dil içeren 3.6 bin saatlik Common Voice ve 17 dil içeren 1.7 bin saat uzunluğundaki Babel veri kümesiyle ön eğitilmiştir. 53 dilde 56.000 saat etiketsiz veriyle ön eğitilmiş bu model ile birlikte çok sayıda dilde konuşma tanıma olanağı sunulmuştur. 2021 yılında ise 128 farklı dilde 436.000 saat eğitim verisi kullanılarak XLSR-53 modeli üzerinde önemli iyileştirmeler gerçekleştirilmiş ve Wav2Vec2-XLS-R modeli yayınlanmıştır [24]. XLSR-53 modelini temel alan bu model ile daha fazla dil içeren daha büyük miktarda etiketlenmemiş eğitim verisi (Multilingual LibriSpeech, Common Voice, Voxlingua107, Babel, VoxPopuli) kullanarak gelişmiş bir model oluşturmayı amaçlamışlardır. Çizelge-2'de eğitilebilir parametre sayısına göre Wav2Vec2-XLSR modelleri verilmiştir. Çizelgede M: Milyon adet parametre sayısını, B: Bin saat eğitim verisi süresini ifade etmektedir. Parametrelere dayalı eğitim süresi göz önüne alındığında Wav2Vec2-XLS-R-2B, 2 milyar parametre ile en büyük ve yavaş modelken 300M en küçük ve hızlı modeldir. Donanım ve zaman düşünüldüğünde bu çalışmada 300M modelinin kullanılmasına karar verilmiştir.

**Çizelge-2: Wav2Vec2-XLS-R modellerinin özellikleri tablosu**

| Model | Parametre | Süre (saat) | Çok Dilli |
|---|---|---|---|
| **Wav2Vec2-XLS-R-300M** | **300M** | **436B** | ✓ |
| Wav2Vec2-XLS-R-1B | 1 milyar | 436B | ✓ |
| Wav2Vec2-XLS-R-2B | 2 milyar | 436B | ✓ |

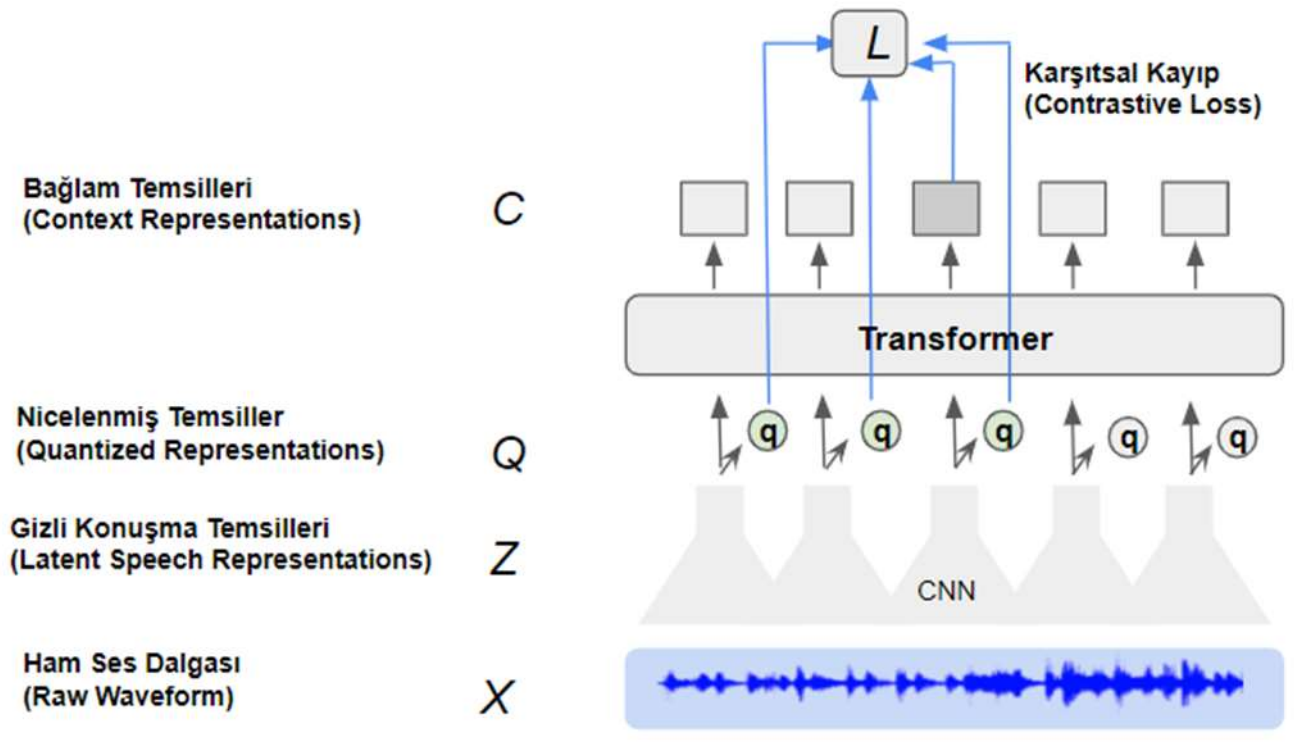

**Şekil-2:** Wav2Vec2.0 mimarisi [18]

Wav2vec2.0 modeli, konuşmayı çok katmanlı bir evrişimli sinir ağı aracılığıyla kodlar ve ardından maskelenmiş dil modellemeye benzer şekilde ortaya çıkan gizli konuşma temsillerinin yayılma alanlarını maskeler (Şekil-2). Gizli temsiller, bağlamsallaştırılmış temsiller oluşturmak için bir dönüştürücü ağına beslenir ve model çeldiricilerden ayırt edeceği karşılaştırmalı bir görevle eğitilir. Karşılaştırmalı görevdeki gizli temsilleri temsil etmek için ayrık konuşma birimlerini öğrenir. Etiketlenmemiş konuşmaya ilişkin ön eğitimden sonra model, konuşma tanıma görevleri için kullanılmak üzere etiketli veriler ile ince ayar yapılır.

### 3.3.2 Whisper-Small

Whisper, OpenAI tarafından geliştirilen ve konuşmaları, konuşulduğu dilde metne dönüştüren veya eğitildiği herhangi başka bir dile çevirebilen bir otomatik konuşma tanıma (ASR) sistemidir [19]. Sistem, aksanlara, arka plan gürültüsüne ve teknik dile karşı sağlamlık sağlayan web'den toplanan 680.000 saatlik çok dilli ve çok görevli denetimli veriden oluşan geniş ve çeşitli bir veri kümesi üzerinde eğitilmiştir [19]. Çizelge-3'de Whisper modelinin farklı büyüklüklerdeki özellikleri gösterilmektedir. M: Milyon adet parametre sayısını ifade etmektedir [25].

**Çizelge-3: Whisper modellerinin özellikleri tablosu [25]**

| Boyut | Parametre | İngilizce | Çok Dilli |
|---|---|---|---|
| Tiny | 39 M | ✓ | ✓ |
| Base | 74M | ✓ | ✓ |
| **Small** | **244M** | ✓ | ✓ |
| Medium | 769M | ✓ | ✓ |
| Large | 1550M | | ✓ |

Whisper-Small modeli, 244 milyon parametre ile geliştirilen dokuz Whisper modellinden biridir. Orta büyüklükte olması ve kapasite olarak Wav2Vec2.0'da kullandığımız modele benzemesi nedeniyle bu çalışma için seçilmiştir.

Whisper mimarisi, bir kodlayıcı-kod çözücü Transformatörü olarak uygulanan uçtan uca basit bir yaklaşımdır. Giriş sesi 30 saniyelik parçalara bölünür ve bir log-Mel spektrogramına dönüştürülür. Spektrogram daha sonra bir kodlayıcıya iletilir ve bir kod çözücü karşılık gelen metin başlığını tahmin etmesi için eğitilir. Kod çözücü ayrıca, modeli dil tanımlama, tümce düzeyinde zaman damgaları, çok dilli konuşma transkripsiyonu ve konuşma çevirisi gibi görevleri gerçekleştirmeye yönlendiren özel belirteçleri karıştırır [19].

Şekil-3'de gösterildiği gibi, Whisper mimarisi bir kodlayıcı ve bir kod çözücüden oluşur. Kodlayıcı, sesin log-Mel spektrogramını girdi olarak alır ve daha sonra kod çözücüye iletilen gizli bir temsili çıkarır. Kod çözücü, nihai transkripsiyonu veya çeviri çıktısını tahmin eder.

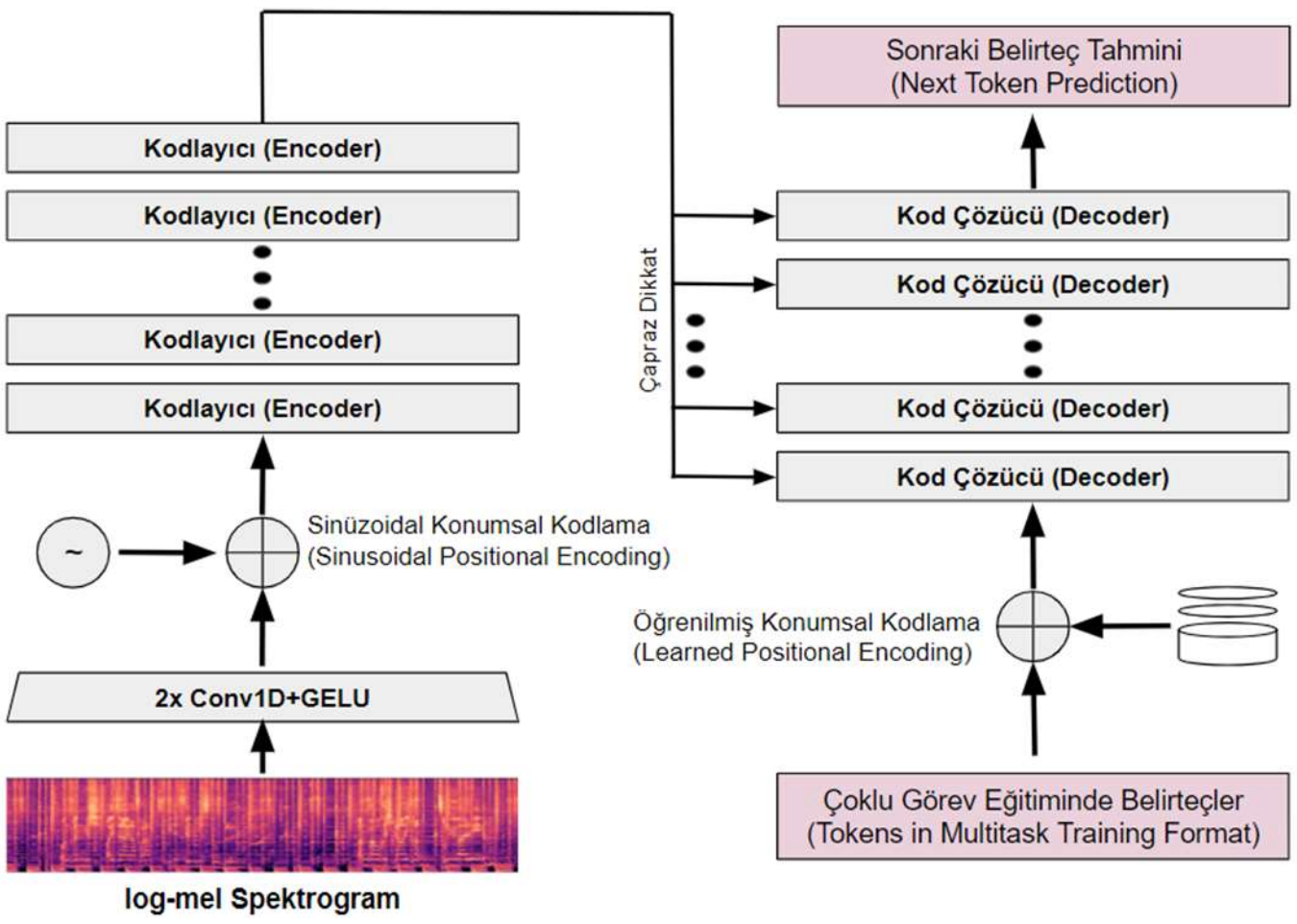

**Şekil-3:** Whisper mimarisi [19]

### 3.3.3 Başarı Metrikleri

Sesten metne çeviri modellerinin başarılarını değerlendirmek için başarı metrikleri kullanılmıştır. Model performanslarının analizi için popüler teknikler olan kelime hata oranı ve karakter hata oranı bu bölümde açıklanmıştır.

#### 3.3.3.1 Kelime Hata Oranı

Kelime hata oranı (Word Error Rate (WER)) konuşma tanıma sisteminin performansını analiz etmek için kullanılan popüler bir tekniktir [26]. WER hesaplaması Levenshtein mesafesini temel alır. Levenshtein mesafesi iki kelime arasındaki farklılığı harf düzeyinde analiz eder. WER ise çeviri performansını kelime düzeyinde analiz etmekte olup sesten metne çeviri başarısı, sistem çıktısı ile referans kelimelerin eşleşmesine dayanmaktadır. WER değeri insan tarafından yazıya çevrilmiş referans kelime dizisi dikkate alındığında sesten metne çeviri dizisinde gözlemlenen silinen kelime, eklenen kelime ve değişen kelime sayısına göre hesaplanır. Eklenen kelime sayısı I, silinen kelime sayısı D, değişen kelime sayısı S olmak üzere toplam hatanın referans kelime dizideki kelime sayısı N'e oranıyla WER değeri hesaplanır. (Denklem 1)

$$WER = \frac{I+D+S}{N} \quad (1)$$

#### 3.3.3.2 Karakter Hata Oranı

Karakter hata oranı (Character Error Rate (CER)), WER ile benzerlik göstermekte olup farklı olarak doğruluk karakter düzeyinde analiz edilmektedir [27]. Bu metrik ile yanlış tahmin edilen karakterlerin oranı elde edilmektedir. CER değeri ne kadar düşükse, konuşma tanıma sistemi o kadar başarılı demekle beraber '0' en iyi puandır. Eklenen karakter sayısı I, silinen karakter sayısı D, değişen karakter sayısı S, referans metindeki karakter sayısı N olmak üzere CER değeri Denklem 2'deki gibi hesaplanır.

$$CER = \frac{I+D+S}{N} \quad (2)$$

## 4. Deneysel Sonuçlar

Bölüm 2’de model eğitimleri ve doğrulamasında kullanılacak veri kümesi, veri kümesinde uygulanan ön işlemler, sesten metne çeviri için önerilen derin öğrenme mimarileri ve bu mimarilerin performans değerlendirmesinde kullanılan başarı metrikleri açıklanmıştır. Bu bölümde ise model eğitimleri sonucunda elde edilen veriler sunulacaktır.

Çalışmada Türkçe dilinde konuşmadan metne çeviri gerçekleştirmek amaçlamış olup bu doğrultuda literatürde yer alan iki farklı model eğitilerek sonuçları değerlendirilmiştir. Modellerin eğitimi ve doğrulamasında Common Voice Corpus 11.0 kullanılmıştır. Veri kümesindeki 36125 adet veriyle model eğitimi, 10143 adet veriyle modelin doğrulaması gerçekleştirilmiştir. Her iki model için eğitimden önce veri kümesi ön işlemlerden geçirilmiş veri üzerinde bir takım düzenlemeler gerçekleştirilmiştir (Şekil-4). Kullandığımız veri kümesindeki metinler herhangi bir ses birimine karşılık gelmeyen noktalama işaretleri içermektedir. Wav2Vec2-XLS-R-300M modeline uygun olarak veri kümesindeki metinler noktalama işaretlerinden arındırılmıştır. Ek olarak Wav2Vec2-XLS-R-300M modeli veri kümesindeki her harf için bir belirtecin atandığı sözlük oluşturmayı gerektirmektedir. Bu nedenle veri kümesinde yer alan her harf tespit edilip her birine bir tam sayı atanmıştır. Harflere atanan belirteç haricinde kelimeleri ayırmak için ‘|’, modelin karşılaşabileceği, veri kümesinde yer almayan karakterler için ‘UNK’ ve kelime içerisinde tekrar eden harfleri tespit etmek için ‘PAD’ belirteçleri de tanımlanarak sözlüğü dahil edilmiştir. Ayrıca Common Voice 11.0 veri kümesinde MP3 formatındaki depolanmış kayıtların örnekleme frekansı model girişine uygun olarak 16 kHz'e dönüştürülmüştür.

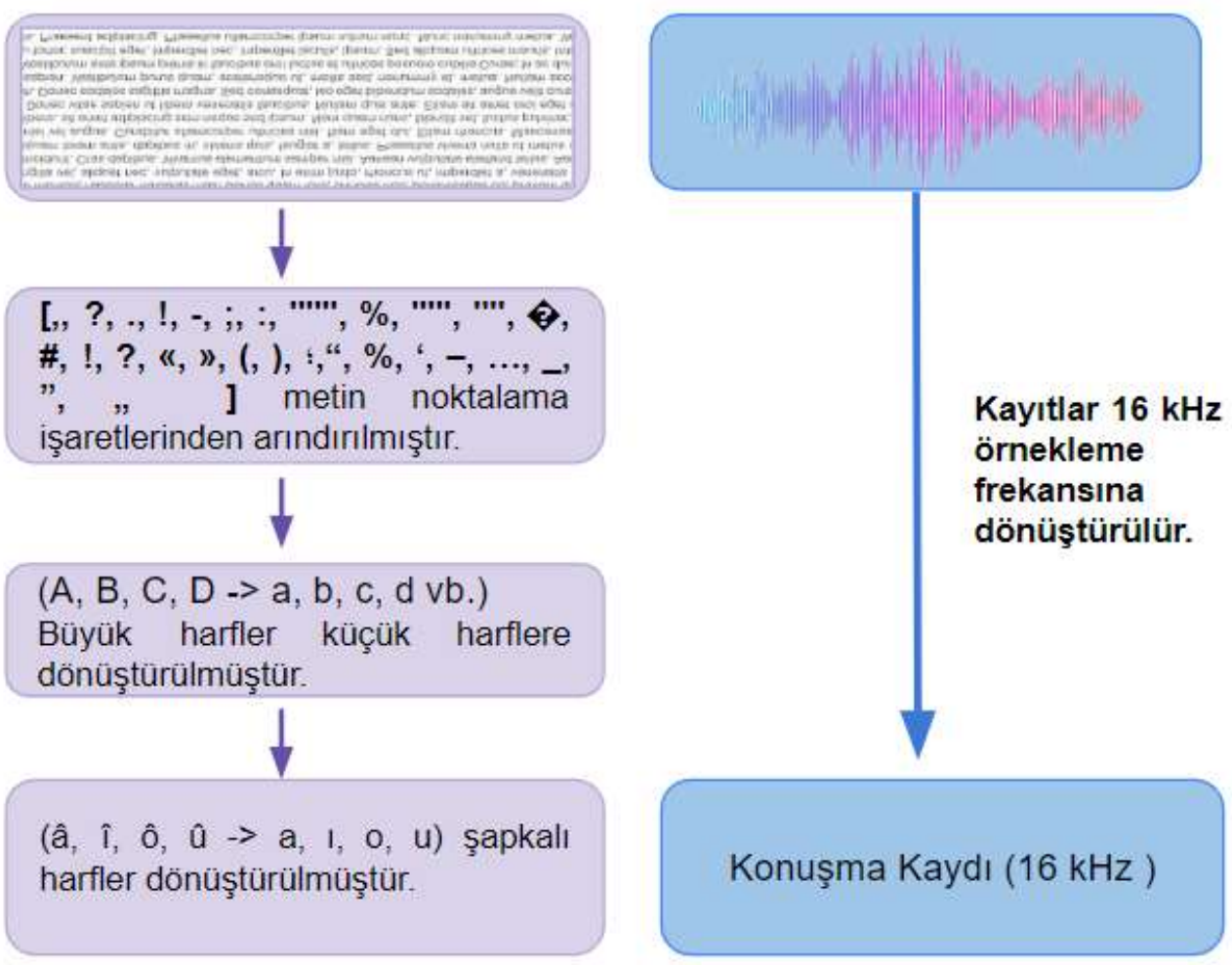

**Şekil-4:** Veri ön işleme akış diyagramı

Whisper-Small model eğitim parametreleri, adım sayısı (step) 5000, öğrenme oranı (learning rate) 1e-05, batch boyutu 32 ve Adam optimizasyon algoritması olarak belirlenmiştir. Wav2Vec2-XLS-R-300M model eğitim parametreleri ise eğitim tur sayısı (epoch) 30, öğrenme oranı (learning rate) 3e-04, batch boyutu 64 ve Adam optimizasyon algoritması olarak belirlenmiştir. Her iki model için farklı parametrelerle çok sayıda deney gerçekleştirilmiş bu deneyler sonucunda en yüksek başarıların elde edildiği parametreler model parametreleri olarak belirlenmiştir.

Şekil-5 ve Şekil-6’da sırasıyla ince ayarlanmış Wav2Vec2-XLS-R-300M [28] ve Whisper-Small [29] modellerinin adım sayısı boyunca gözlemlenen kayıp grafikleri verilmiştir. Şekil-6’da Whisper-Small modelinin 2.000’inci adım sayısında en düşük doğrulama kaybına ulaştığı sonrasında aşırı öğrenme (overfitting) gösterdiği gözlemlenmiş ve grafikte siyah nokta ile belirtilmiştir. Şekil-5’de ise Wav2Vec2-XLS-R-300 modeli için böyle bir durumun söz konusu olmadığı kayıp grafiklerinde düşüşün devam ettiği gözlemlenmektedir. Bu nedenle Wav2Vec2-XLS-R-300M modelinin eğitimi devam ettirilmiştir. Wav2Vec2-XLS-R-300M ve Whisper-Small modelleri için sırasıyla Şekil-7 (a) ve Şekil-7 (b)’de eğitim ve doğrulama kayıp grafikleri tek bir grafikte birleştirilerek modellerin eğitim süreci daha anlaşılır kılınmıştır. Modellerin STT başarısını değerlendirmek amacıyla eğitim süresince en düşük WER değerinin gözlemlendiği adım sayısındaki model kaydedilmiştir. Çizelge-4’te modellerin en düşük WER değerinin elde edildiği adım sayısındaki eğitim, doğrulama kayıpları ve STT başarısını değerlendirmek için kullandığımız iki performans metriği olan WER ve CER değeri verilmiştir. Elde edilen sonuçlar incelendiğinde Whisper-Small modelinin başarılı sonuçlar verdiği gözlemlenmiştir. Ayrıca eğitim süresi göz önüne alındığında Whisper-Small model eğitimi daha az adım sayısında başarılı sonuçlar vermiş olup Wav2Vec2-XLS-R-300M modeline kıyasla eğitim daha kısa sürede sonuçlanmıştır.

**Çizelge-4: Wav2Vec2-XLS-R-300M ve Whisper-Small modelleri sonuç tablosu**

| Model | Eğitim Kaybı | Doğrulama Kaybı | WER | CER | Adım Sayısı |
|---|---|---|---|---|---|
| **Wav2Vec-xls-r-300m** | 0,06 | 0,31 | 0,28 | 0,06 | 16800 |
| **Whisper-Small** | 0,001 | 0,25 | 0,16 | 0,04 | 4000 |

Modellerin performansını değerlendirmek için eğitim ve doğrulama veri kümesinde yer almayan kayıtlarla modeller test edilmiştir. Test verisi hazırlanırken AdresGezgini A.Ş. veri tabanında kayıtlı çağrı merkezi kaydı örneklerinden yararlanılmıştır. Farklı kişilerin konuştuğu çağrı kayıtlarından seçilen ve yaklaşık 3 saat uzunluğundaki test verisi yazıya çevrilerek her ses kaydına ait etiket oluşturulmuştur. Test veri kümesindeki her bir kayıt Common Voice veri kümesiyle en iyileştirmesini gerçekleştirdiğimiz Whisper-Small ve Wav2Vec2-XLS-R-300M modelleriyle sesten metne çevrilmiştir. İnsan tarafından etiketlendirilmiş çağrı merkezi kayıtlarının konuşmadan metne çevirisi referans metin olarak kabul edilmiş ve model çıktılarıyla referans metinler kıyaslanarak modellerin test kayıtları üzerindeki WER değeri hesaplanmıştır. Test verileriyle gerçekleştirilen bu değerlendirme sonucunda Whisper-Small modelinin Wav2Vec2-XLS-R-300M modeline göre sesten metne çeviride daha yüksek başarı gösterdiği gözlemlenmiştir. Ancak Şekil-8'de de görüldüğü üzere çağrı merkezi kayıtlarıyla gerçekleştirilen testte her iki model de yüksek WER değerlerine sahiptir. Çağrı merkezi kayıtlarının net şekilde anlaşılmayan, temiz kayıtlar olmaması ayrıca eğitim veri kümesinde çağrı merkezi kayıtlarına hiç yer verilmemesi her iki modelin de bu kayıtlardaki başarısını düşürmüştür.

Ayrıca Wav2Vec2-XLS-R-300M modelinin STT çıktısı noktalama işaretleri ve küçük büyük harf uyumu bulunmayan metinler iken Whisper-Small modeli farklı olarak metin çıktısına noktalama işaretlerini eklemektedir. Fakat bu çalışmada modelleri aynı özellikteki birebir aynı veri kümesiyle ince ayar yaparak başarılarını analiz etmek amacıyla her iki model için de veri kümesi aynı ön işlemlerden geçirilmiş ve eğitim veri kümesi noktalama işaretleri, özel karakterler, büyük harflerden ve şapkalı harflerden arındırılmıştır. Bu nedenle bu çalışma sonucunda elde edilen ince ayarlanmış Whisper-Small modelinin konuşmadan metne çevirisi noktalama işaretleri içermemektedir. Bununla birlikte Şekil-8'de görüleceği üzere Whisper-Small modelinin genelleme yetisi seçili Wav2Vec2-XLS-R-300M modeline göre belirli ölçüde daha iyidir. Son olarak Çizelge-5'te eğitim ve doğrulama veri setine dahil edilmemiş Common Voice verisinden alınmış ses kayıtlarının konuşmadan metne çıktıkları verilmiştir. Referans metin ve model çıktıları karşılaştırıldığında Whisper-Small modelinin daha başarılı sonuçlar verdiği gözlemlenmektedir.

Önerilen çalışma Python programlama dili ile gerçekleştirilmiştir. Verilerin düzenlenmesi için Numpy ve Pandas kütüphanelerinden faydalanılmıştır. Eğitim sonuçlarının gösterilmesi için TensorBoard kullanılmıştır. Model eğitimleri 12 çekirdekli 2.90 GHz Intel(R) i5-10400F CPU, 48 GB 3600MHz DDR4 RAM ve GeForce RTX 3060 12 GB GPU'ya sahip cihazda gerçekleştirilmiştir.

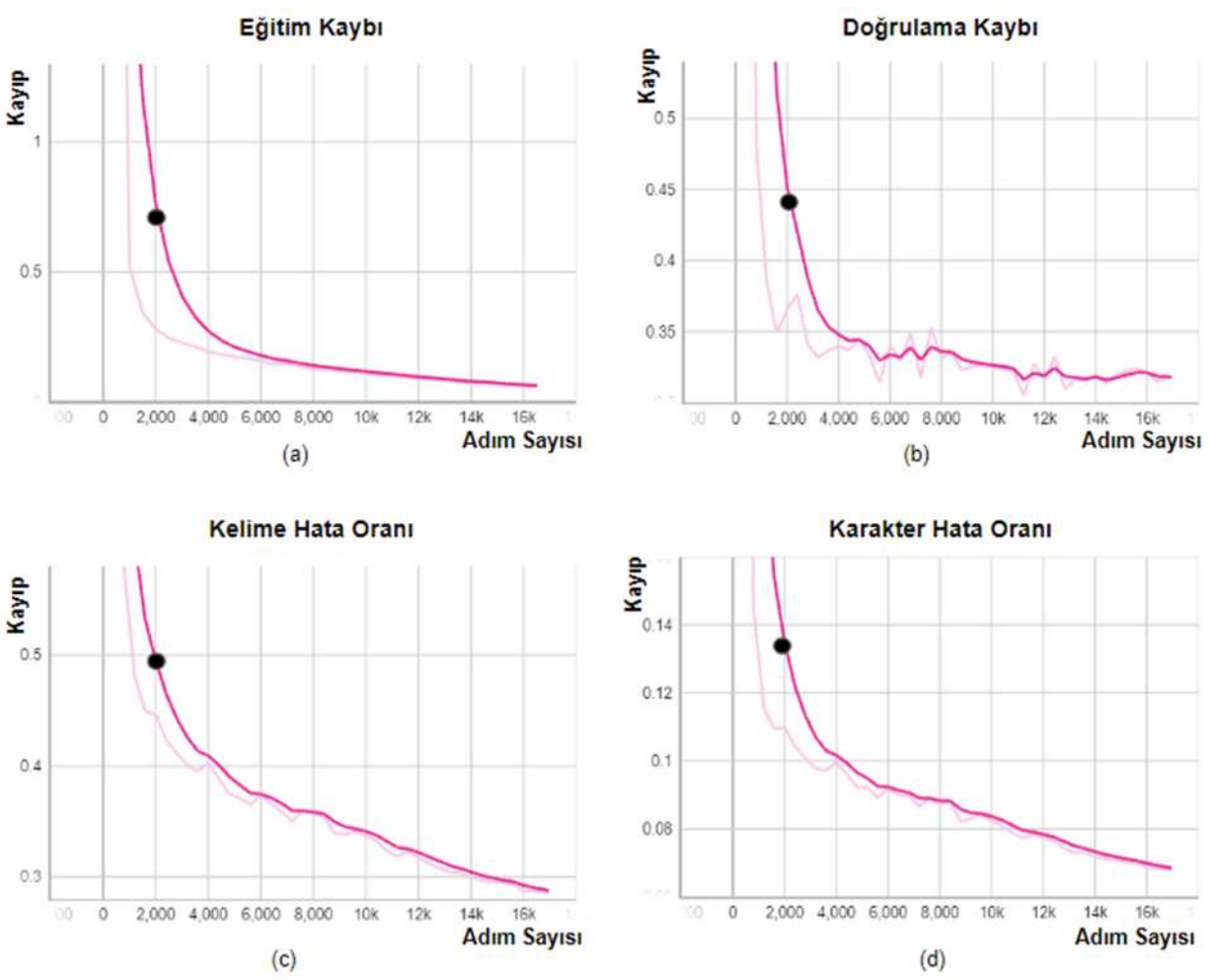

**Şekil-5:** Wav2Vec2-XLS-R-300M eğitim kaybı (a), doğrulama kaybı (b), model başarısını kelime düzeyinde analiz eden kelime hata kaybı (c), model başarısını karakter düzeyinde analiz eden karakter hata kaybı (d) grafikleri

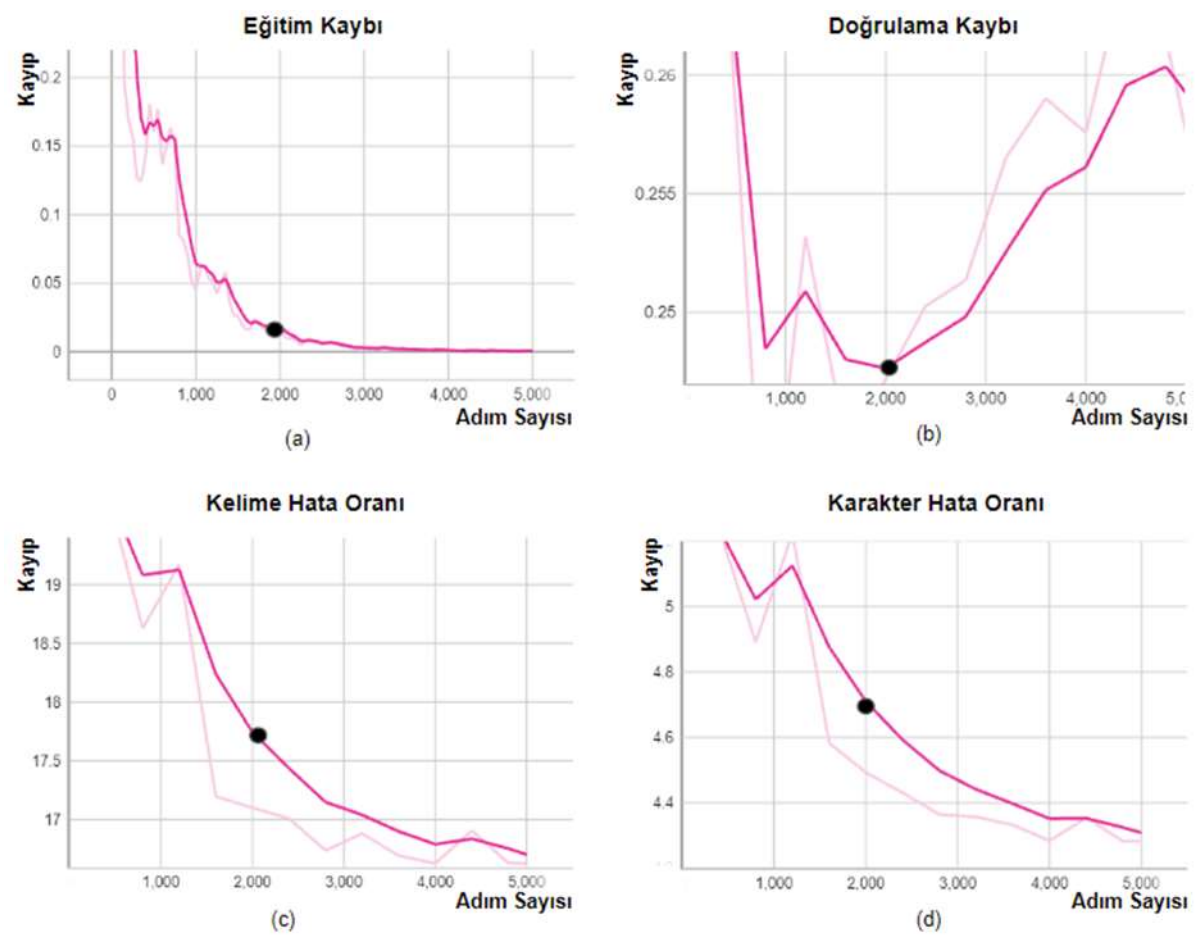

**Şekil-6:** Whisper-Small eğitim kaybı (a), doğrulama kaybı (b), model başarısını kelime düzeyinde analiz eden kelime hata kaybı (c), model başarısını karakter düzeyinde analiz eden karakter hata kaybı (d) grafikleri

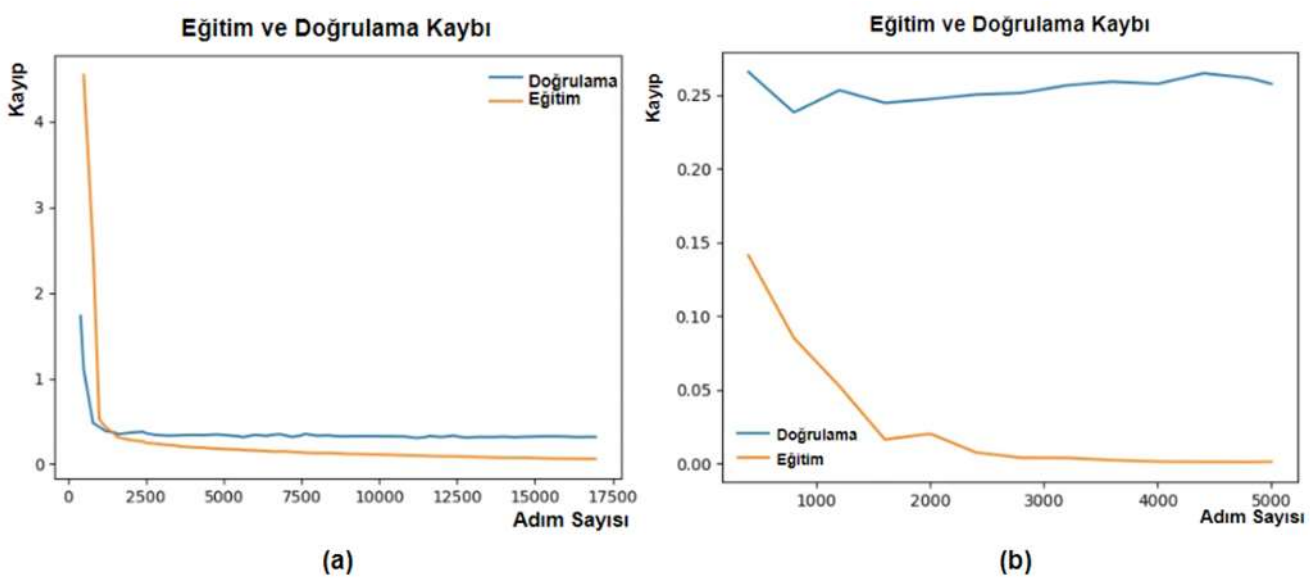

**Şekil-7:** Wav2Vec2-XLS-R-300M modeli eğitim kaybı ve doğrulama kaybı (a), Whisper-Small modeli eğitim kaybı ve doğrulama kaybı (b) grafikleri

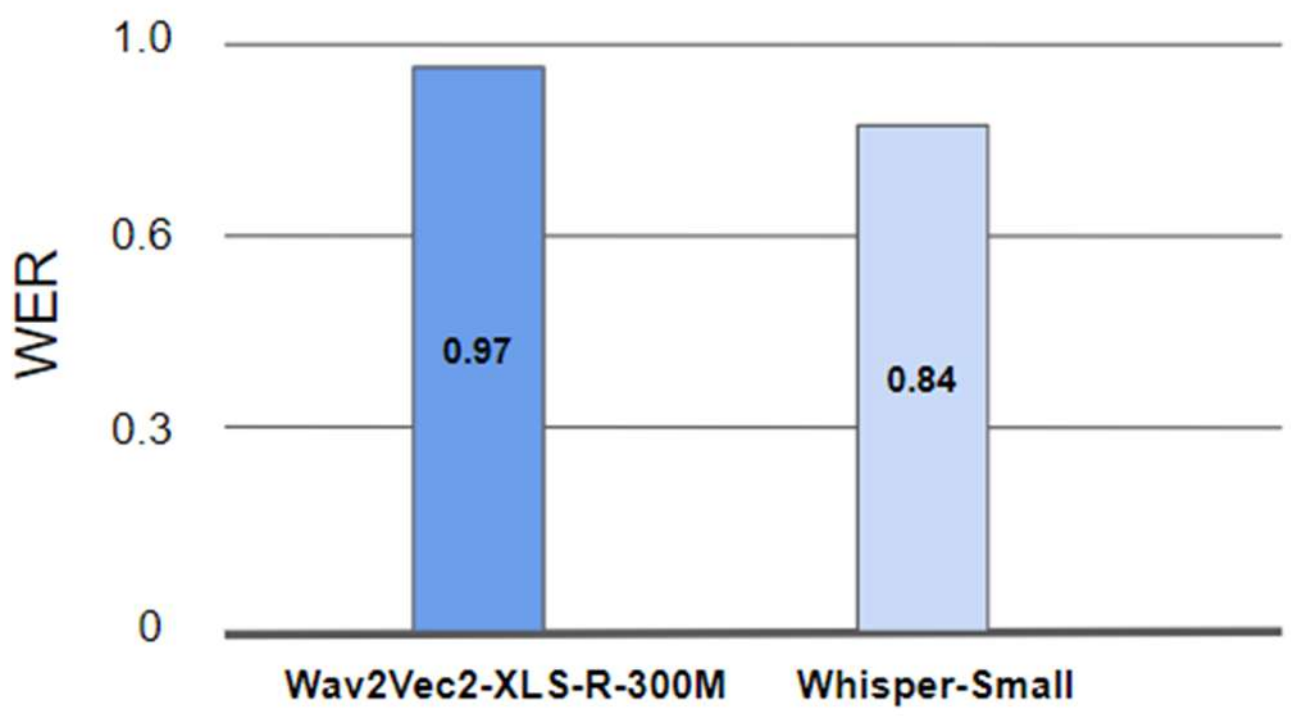

**Şekil-8:** Wav2Vec2-XLS-R-300M ve Whisper-Small modellerinin çağrı merkezi kayıtlarında WER değerleri

**Çizelge-5: Referans metin ile birlikte ince ayarlanmış Wav2Vec2-XLS-R-300M ve Whisper-Small modelleriyle konuşmadan metne çeviri örnekleri**

| Referans Metin | Wav2Vec2-XLS-R-300M | Whisper-Small |
|---|---|---|
| Son birkaç yılda nasıl bir değişim oldu | son birkaç yılda nasıl bir değişim olduk | son birkaç yılda nasıl bir değişim olduk |
| evrak işlerini halletmek iki üç ay sürerse yunanistana hangi şirket gelir | evrak işlerini halletmek iki yüç ay sürerse yunanistan ahangi şirket gelir | evrak işlerini halletmek iki üç ay sürerse yunanistanı hangi şirket gelir |
| güneydoğu avrupa ülkeleri biyoyakıt konusunu değerlendiriyor | güneydoğ ağrup avülkeleri biryoyakıt konusunu değerlendiriyor | güneydoğu avrupa ülkeleri bioyakıt konusunu değerlendiriyor |
| yine teknolojiye güvenip olanları kayıtsız kalmak mümkün mü | yine teknolojiye güvenip olamları kayıfsız kalnmak mümküm mü | yine teknolojiye güvenip olanları kayıtsız kalmak mümkün mü |
| bu bizi her koşulda birleştiren bir şey | bu bizi her koşulda birleştirembi ir şey | bu bizi her koşulda birleştiren bir şey |

## 5. Sonuçlar

Önerilen bu çalışma ile açık kaynaklı bir veri kümesi olan Türkçe dilinde hazırlanmış Common Voice 11.0 ile literatürde yer alan Whisper-Small ve Wav2Vec2-XLS-R-300M STT modelleri ince ayar yapılmıştır. İnce ayar ardından her iki modelin Türkçe dilinde sesten metne çeviri problemi üzerindeki başarıları incelenmiş ve modellerin performansı değerlendirilmiştir. Çalışmada Whisper-Small modelinin 2000. adım sayısında aşırı öğrenme gösterdiği dikkate alınarak Wav2Vec2-XLS-R-300M modelinin de aynı adım sayısındaki performansı incelenmiş fakat aşırı öğrenme gözlenmemesi nedeniyle eğitim devam ettirilmiştir. Her iki model için de en düşük doğrulama kaybının elde edildiği adım sayısındaki modeller kaydedilmiştir. En düşük WER değerinin elde edildiği adım sayısında Whisper-Small modelinde 0,16 WER değeri elde edilirken Wav2Vec2-XLS-R-300M modelinde WER değeri 0,28 olarak elde edilmiştir. Ayrıca şirket bünyesinde gerçekleştirilmiş çağrı merkezi kayıtlarıyla oluşturulmuş test verisiyle modeller test edilmiştir. Eğitim veri kümesinde çağrı merkezi kaydına yer verilmemesi nedeniyle her iki modelde de yüksek WER değeri elde edilmiştir. Ancak yine de gerek daha hızlı bir şekilde eğitilebilmesi, gerekse uygun noktalama işaretlerini üretebilme gerekse genelleme kabiliyetleri açısından Whisper-Small modeli daha tercih edilebilir bir model olarak görünmektedir. Gerçekleştirilen bu çalışmada veri kümesinin tek bir kaynaktan oluşması ve az sayıda veri içerdiği göz önüne alındığında gelecek çalışmalarda veri kaynağı çeşitliliğinin oluşturulması ve çok sayıda veri ile gerçekleştirilecek model ince ayarıyla STT başarısının artırılması hedeflenmektedir.

## 6.Teşekkür

## Kaynaklar

[1] Özlan, B., Haznedaroğlu, A., Arslan, L. M., *Automatic fraud detection in call center conversations,* In 2019 27th Signal Processing and Communications Applications Conference (SIU), 2019, pp. 1-4.

[2] Dhanjal, A. S., Singh, W. *An automatic machine translation system for multi-lingual speech to Indian sign language. multimedia Tools and Applications*, 2022, pp.1-39.

[3] Ballati, F., Corno, F., De Russis, L., Assessing virtual assistant capabilities with Italian dysarthric speech, In Proceedings of the 20th International ACM SIGACCESS Conference on Computers and Accessibility, 2018, pp. 93-101.

[4 ] Hinton, G., Deng, L., Yu, D., Dahl, G. E., Mohamed, A. R., Jaitly, N., Kingsbury, B., *Deep neural networks for acoustic modeling in speech recognition: The shared views of four research Groups,* IEEE Signal processing magazine, 2012, 29(6), pp.82-97.

[5] Sainath, T. N., Vinyals, O., Senior, A., Sak, H. *Convolutional, long short-term memory, fully connected deep neural networks*, IEEE international conference on acoustics, speech and signal processing (ICASSP), 2015, pp. 4580-4584.

[6] Alharbi, S., Alrazgan, M., Alrashed, A., Alnomasi, T., Almojel, R., Alharbi, R., Almojil, M*., Automatic speech recognition: Systematic literature Review*, IEEE Access, 9, 2021, pp.131858-131876.

[7] Hellman, E., Nordstrand, M., *Research in methods for achieving secure voice anonymization: Evaluation and improvement of voice anonymization techniques for whistleblowing,* 2022.

[8] Chorowski, J. K., Bahdanau, D., Serdyuk, D., Cho, K., Bengio, Y., *Attention-based models for speech recognition*, Advances in neural information processing systems, 2015.

[9] Bahar, P., Bieschke, T., Ney, H., *A comparative study on end-to-end speech to text translation,* Automatic Speech Recognition and Understanding Workshop (ASRU), 2019, pp. 792-799.

[10]Tang, Y., Pino, J., Wang, C., Ma, X., Genzel, D., *A general multi-task learning framework to leverage text data for speech to text tasks*, In ICASSP IEEE International Conference on Acoustics, Speech and Signal Processing, 2021, pp. 6209-6213.

[11]Tombaloğlu, B., Erdem, H. A., *SVM based speech to text converter for Turkish language*, In 2017 25th Signal Processing and Communications Applications Conference (SIU), 2017, pp. 1-4.

[12] Kimanuka, U. A., & Buyuk, O., *Turkish speech recognition based on deep neural networks*, Süleyman Demirel Üniversitesi Fen Bilimleri Enstitüsü Dergisi, 2018, pp.319-329.

[13]Tombaloğlu, B., Erdem, H., *Deep Learning Based Automatic Speech Recognition for Turkish*, Sakarya University Journal of Science, 2020*,* pp.725-739.

[14]Tombaloğlu, B., & Erdem, H., *Turkish Speech Recognition Techniques and Applications of Recurrent Units (LSTM and GRU),* Gazi University Journal of Science, 2021, pp.1035-1049.

[15]Safaya, A., Erzin, E., *HuBERT-TR: Reviving Turkish Automatic Speech Recognition with Self-supervised Speech Representation Learning*, 2022, arXiv preprint arXiv:2210.07323.

[16] Li, Z., Niehues, J., *Efficient Speech Translation with Pre-trained Models*, 2022, arXiv preprint arXiv:2211.04939.

[17] Baevski, A., Zhou, Y., Mohamed, A., & Auli, M. *Wav2vec 2.0: A framework for self-supervised learning of speech representations,* Advances in Neural Information Processing Systems, 2020, pp. 12449-12460.

[18] Vásquez-Correa, J. C., Álvarez Muniain, A., *Novel Speech Recognition Systems Applied to Forensics within Child Exploitation: Wav2vec2.0 vs. Whisper,* 2023, Sensors, 23(4), 1843.

[19] Radford, A., Kim, J. W., Xu, T., Brockman, G., McLeavey, C., Sutskever, I., *Robust speech recognition via large-scale weak supervision*, 2022, arXiv preprint arXiv:2212.04356.

[20] D.E. Taşar, *An automatic speech recognition system proposal for organizational development*, 782089, Master's thesis, Dokuz Eylul University Management Information Systems, 2023.

[21] Mercan, Ö. B., Özdil, U., Ozan, Ş., *Çok Dilli Sesten Metne Çeviri Modelinin İnce Ayar Yapılarak Türkçe Dilindeki Başarısının Arttırılması Increasing Performance in Turkish by Finetuning of Multilingual Speech-to-Text Model*, 30th Signal Processing and Communications Applications Conference (SIU), 2022, pp. 1-4.

[22] Arduengo, J., Köhn, A., *The Mozilla Common Voice Corpus. In Proceedings of the 19th Annual Conference of the International Speech Communication Association (INTERSPEECH 2019)*, 2019, pp. 1823-1827.

[23] Conneau, A., Baevski, A., Collobert, R., Mohamed, A., & Auli, M., *Unsupervised cross-lingual representation learning for speech recognition*, 2020, *arXiv preprint arXiv:2006.13979*.

[24] Babu, A., Wang, C., Tjandra, A., Lakhotia, K., Xu, Q., Goyal, N., Auli, M., *XLS-R: Self-supervised cross-lingual speech representation learning at scale*, 2021, *arXiv preprint arXiv:2111.09296*.

[25] Openai. (2022, December 9). *Whisper/model-card.md at main· openai/whisper*. GitHub. Retrieved February 5, 2023,fromhttps://github.com/openai/whisper/blob/main/model-card.md

[26] Ali, A., Renals, S., *Word error rate estimation for speech recognition: e-WER*, In Proceedings of the 56th Annual Meeting of the Association for Computational Linguistics, 2018, pp. 20-24.

[27] Maas, A., Xie, Z., Jurafsky, D., & Ng, A. Y., *Lexicon-free conversational speech recognition with neural networks*, In Proceedings of the 2015 Conference of the North American Chapter of the Association for Computational Linguistics: Human Language Technologies, 2015, pp. 345-354.

[28]“wav2vec2-xls-r-300m-tr”,https://huggingface.co/Sercan/wav2vec2-xls-r-300m-tr

[29]“whisper-small-tr-2”,https://huggingface.co/Sercan/whisper-small-tr-2